\documentclass[10pt,twocolumn,letterpaper]{article}

\usepackage{iccv}
\usepackage{times}
\usepackage{epsfig}
\usepackage{graphicx}
\usepackage{amsmath}
\usepackage{amssymb}
\usepackage{graphicx}
\usepackage[table,xcdraw]{xcolor}
\usepackage{amsmath,graphicx}
\usepackage{algorithm}
\usepackage{multirow}
\usepackage{booktabs}
\usepackage[table,xcdraw]{xcolor}
\usepackage[accsupp]{axessibility}
\usepackage{graphicx}
\usepackage{amsmath}
\usepackage{amssymb}
\usepackage{booktabs}
\usepackage{multirow}
\usepackage{amsthm}
\usepackage{caption}
\usepackage{subcaption}
\usepackage{algorithm}
\usepackage{algpseudocode}
\usepackage{amsmath,amssymb,amsfonts}
\usepackage{colortbl}

\usepackage{accents}
\usepackage{amsmath,amssymb,amsfonts}
\usepackage{nicefrac}      
\usepackage{algorithm}
\usepackage{algpseudocode}
\usepackage{gensymb}

\usepackage{bm}

\newcommand{\hs}{BTC }
\newcommand{\hsnospace}{BTC}
\newcommand{\hsfull}{Breaking Temporal Consistency }

\newcommand{\hsuap}{BTC-UAP }
\newcommand{\hsuapnospace}{BTC-UAP}

\usepackage[breaklinks=true,bookmarks=false]{hyperref}

\makeatletter
\DeclareRobustCommand\onedot{\futurelet\@let@token\@onedot}
\def\@onedot{\ifx\@let@token.\else.\null\fi\xspace}

\makeatother

\iccvfinalcopy 

\def\iccvPaperID{10052} 

\ificcvfinal\fi

\begin{document}

\title{Breaking Temporal Consistency: \\Generating Video Universal Adversarial Perturbations Using Image Models} 

\author{Hee-Seon Kim, Minji Son, Minbeom Kim, Myung-Joon Kwon, Changick Kim\\
Korea Advanced Institute of Science and Technology (KAIST)\\
{\tt\small \{hskim98, ming0103, alsqja1754, kwon19, changick\}@kaist.ac.kr}
}

\maketitle
\ificcvfinal\thispagestyle{empty}\fi

\begin{abstract}
As video analysis using deep learning models becomes more widespread, the vulnerability of such models to adversarial attacks is becoming a pressing concern.
In particular, Universal Adversarial Perturbation (UAP) poses a significant threat, as a single perturbation can mislead deep learning models on entire datasets.
We propose a novel video UAP using image data and image model. This enables us to take advantage of the rich image data and image model-based studies available for video applications. 
However, there is a challenge that image models are limited in their ability to analyze the temporal aspects of videos, which is crucial for a successful video attack.
To address this challenge, we introduce the \hsfull(\hsnospace) method, which is the first attempt to incorporate temporal information into video attacks using image models. 
We aim to generate adversarial videos that have opposite patterns to the original. 
Specifically, \hsuap minimizes the feature similarity between neighboring frames in videos.
Our approach is simple but effective at attacking unseen video models. 
Additionally, it is applicable to videos of varying lengths and invariant to temporal shifts.
Our approach surpasses existing methods in terms of effectiveness on various datasets, including ImageNet, UCF-101, and Kinetics-400.
\vspace{-0.3cm}

\end{abstract}

\section{Introduction}

Deep learning models have achieved remarkable performance in various computer vision tasks \cite{deng2019arcface, bojarski2016end, szegedy2016rethinking, huang2017densely, bertasius2021space}, including image and video recognition. 
However, there is growing concern about the robustness and reliability of these models, as they have been shown to be vulnerable to adversarial attacks \cite{szegedy2014intriguing, dong2018boosting, xie2019improving}.
Adversarial attacks use imperceptible perturbations to manipulate the inputs to produce inaccurate predictions.
These attacks can have serious consequences in various applications of deep neural networks, such as autonomous vehicles and surveillance cameras \cite{thys2019fooling} where false activity detection \cite{kong2020physgan} can cause serious consequences.
Despite these concerns, the problem of adversarial attacks on video models remains largely unsolved.

\begin{figure}[t!]
     \centering
         \centering
         \includegraphics[width=0.475 \textwidth,trim={0cm 0cm 0cm 0cm},clip]{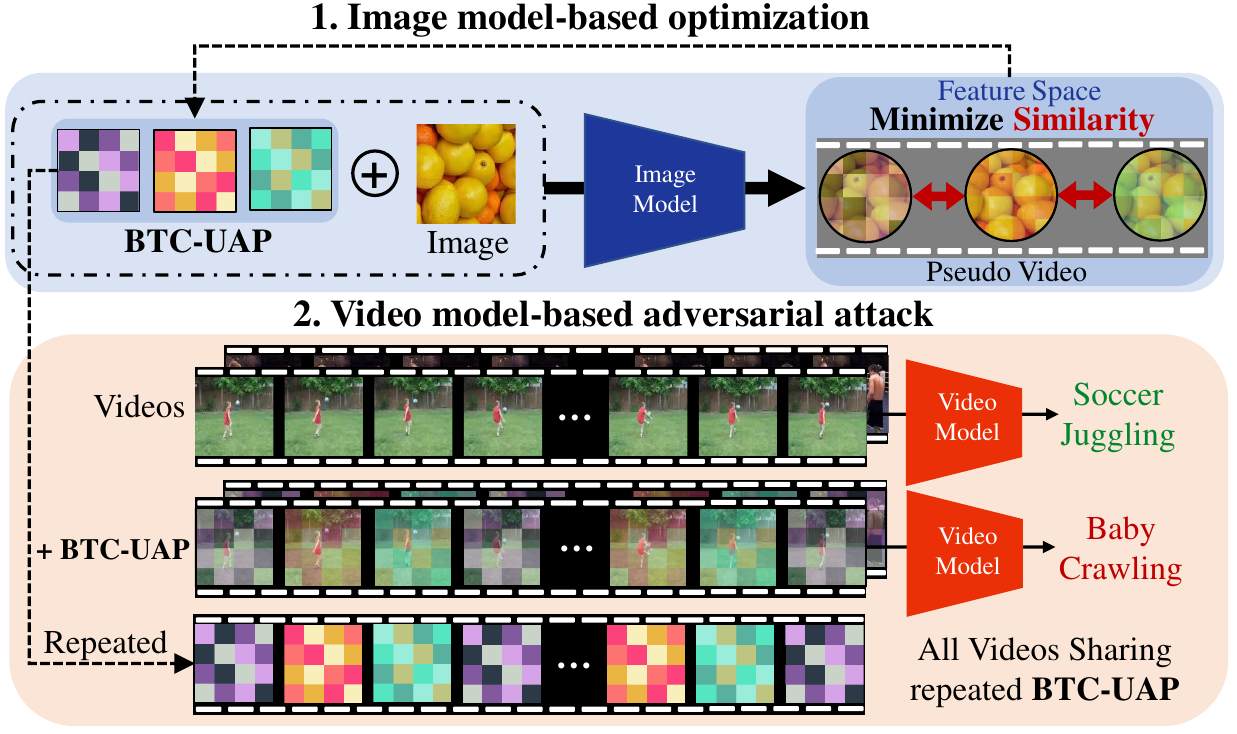}
         \vspace{-0.5cm}
         \caption{
        \textbf{Overall illustration of Breaking Temporal Consistency Method.} 
We propose a novel approach to minimize the similarity between features of consecutive frames in video adversarial attacks. Please note that the illustrated \hsuap is not a real representation, but rather serves as a visual aid. The different colors represent the low similarity between features. 
        }
        \label{fig:fig0}
        \vspace{-0.5cm}
\end{figure}

\begin{figure*}[t!]
     \centering
         \centering
         \includegraphics[width=0.98 \textwidth,trim={0cm 0cm 0cm 0cm},clip]{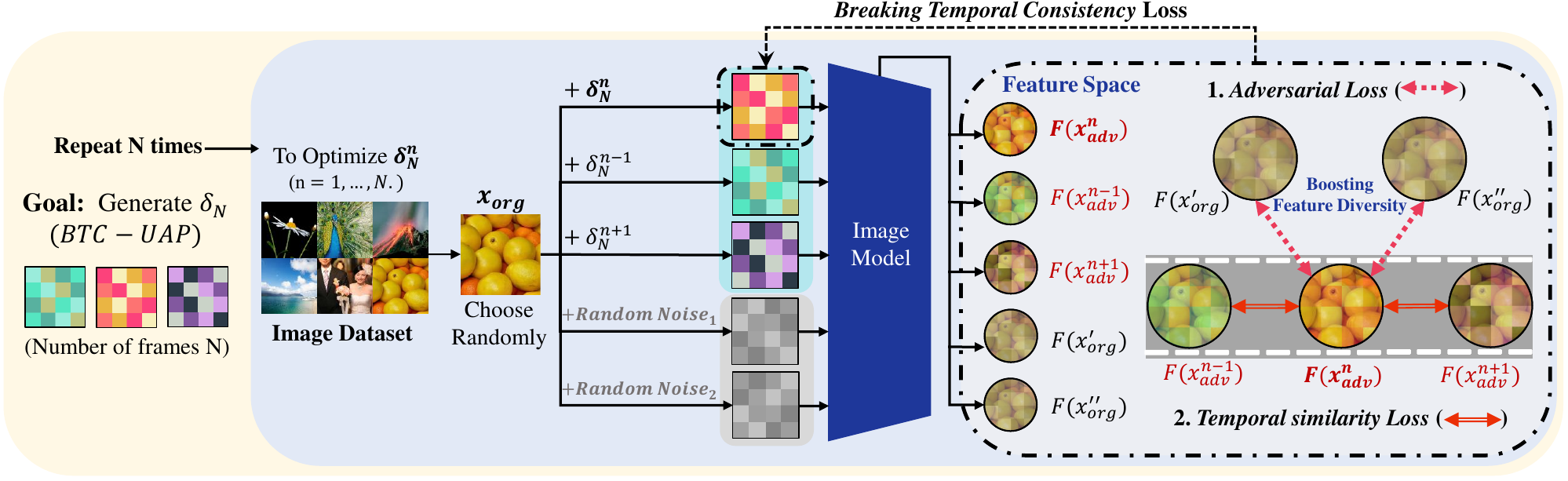}
         \vspace{-0.0cm}
         \caption{
        \textbf{Details of \hsfull Method.} 
Our goal is to create \hsuap for video attacks composed of N frames. We treat each frame of the UAP as an individual image, and add it to the original image to generate corresponding adversarial images. To ensure that these images are adversarial, we use an Adversarial Loss and prevent overfitting with the Feature Diversity method. Additionally, while treating the adversarial images as a pseudo video, we apply the Temporal Similarity Loss to the video frames and make each frame distinct from one another.
        }
        \label{fig:fig1}
        \vspace{-0.3cm}
\end{figure*}

Adversarial attacks can be broadly categorized into white-box \cite{wei2019sparse, hwang2021just, pony2021over} and black-box \cite{TTwei2022boosting, I2Vwei2022cross} attacks. 
White-box attacks exploit model information to generate adversarial examples, while black-box attacks are more challenging due to the lack of model access.
In real-world scenarios, accessing the target model is often difficult or impossible, so black-box attacks are more practical.
One way to launch black-box attacks is by leveraging the transferability of adversarial examples \cite{TTwei2022boosting, I2Vwei2022cross, xie2019improving, dong2019evading, lin2020nesterov}, applying adversarial examples crafted using accessible source models to the target models.
Transfer-based attacks can also be cross-modal \cite{I2Vwei2022cross}, which enables attackers to transfer adversarial examples between different modalities, such as image to video.
For most cases, crafting adversarial examples still requires optimization for each individual adversarial example.
On the other hand, Universal Adversarial Perturbations (UAPs) \cite{moosavi2017universal, wei2019sparse, hwang2021just, xie2022universal, li2018stealthy} poses a powerful threat as a single perturbation can mislead deep learning models on entire datasets. 
This is considered a highly practical attack method in scenarios where it may be difficult or impossible to optimize adversarial perturbations for each individual dataset every time, such as real-time systems.

Our study aims to extend the applicability of UAPs generated using image data and models, to the domain of video data and models. The overall scheme is illustrated in Fig. \ref{fig:fig0}.
This extension allows significant benefits as it allows us to leverage the wealth of image data \cite{deng2009imagenet} and image model-based studies \cite{dong2018boosting, xie2019improving, dong2019evading, lin2020nesterov, byun2022improving} available for video applications. 
Furthermore, generating UAPs using image data requires relatively less computation compared to using video data.
However, we face significant challenges due to the lack of access to video data \cite{soomro2012ucf101, kay2017kinetics} and video models \cite{feichtenhofer2019slowfast, yang2020temporal, wang2018non}.
There are two main challenges in generating adversarial videos using image models only \cite{he2016deep, simonyan2014very, iandola2016squeezenet}. 
Firstly, image models have limited capability in effectively analyzing the passage of time, which is a crucial aspect for videos.
Secondly, UAPs should be applicable to unseen videos of varying lengths. 
Despite the importance of temporal information, prior research has not been able to address these challenges.

As the first paper to consider temporal information in video attacks using image models and data, our study addresses this issue with the \textbf{\hsfull (\hsnospace)} method, as illustrated in Fig. \ref{fig:fig1}. Our target UAP is a video consisting of N frames.
Motivated by the high similarity pattern between neighboring frames in the original video, our UAP aims to generate adversarial videos that have opposite patterns to the original. 
To achieve this, we jointly optimize the adversarial and temporal aspects of the UAPs.
First, to make the UAPs adversarial, we minimize the feature similarity between the original and adversarial images in the feature space using the \textit{Adversarial Loss}.
We treat the frames of the UAPs as images, and add them to the original to create corresponding adversarial images. 
To ensure universality across unseen datasets and prevent overfitting, we incorporate randomness using the Feature Diversity method.
Second, we minimize the similarity between each frame of the UAPs using the \textit{Temporal Similarity Loss}.
To achieve this, we treat the adversarial images as a pseudo-video sequence and minimize the similarity among them.

We named our proposed UAP as \hsuapnospace, which stands for \hsfull Universal Adversarial Perturbation.
To ensure length-agnosticity of \hsuapnospace, we apply it repeatedly until it covers entire frames of the video.
Moreover, our approach is temporal shift invariant, meaning that the starting point of the UAP is irrelevant.
Through extensive experiments on various datasets, including ImageNet, UCF-101, and Kinetics-400, we demonstrate that our simple but effective approach achieves superior performance compared to existing methods.

To summarize our study:
\begin{itemize}

\item We propose a novel video UAP using image data and image models, which allows us to leverage the wealth of image data and image model-based studies available for video applications. 

\item Our study proposes the \hsfull method as the first attempt to incorporate temporal information into video attacks using image models. Our \hsuap makes adversarial videos with opposite patterns to the original by minimizing the feature similarity between neighboring frames in videos. 

\item \hsuap is both temporal shift invariant and length-agnostic, making it a highly practical video attack method that can be applied to videos of varying lengths and datasets. We demonstrate the effectiveness of \hsuap through extensive experiments on various datasets, including ImageNet, UCF-101, and Kinetics-400, outperforming existing methods.

\end{itemize}

\begin{figure}[!t]
     \centering
         \centering
         \includegraphics[width=0.54 \textwidth,trim={0cm 0cm 0cm 0cm},clip]{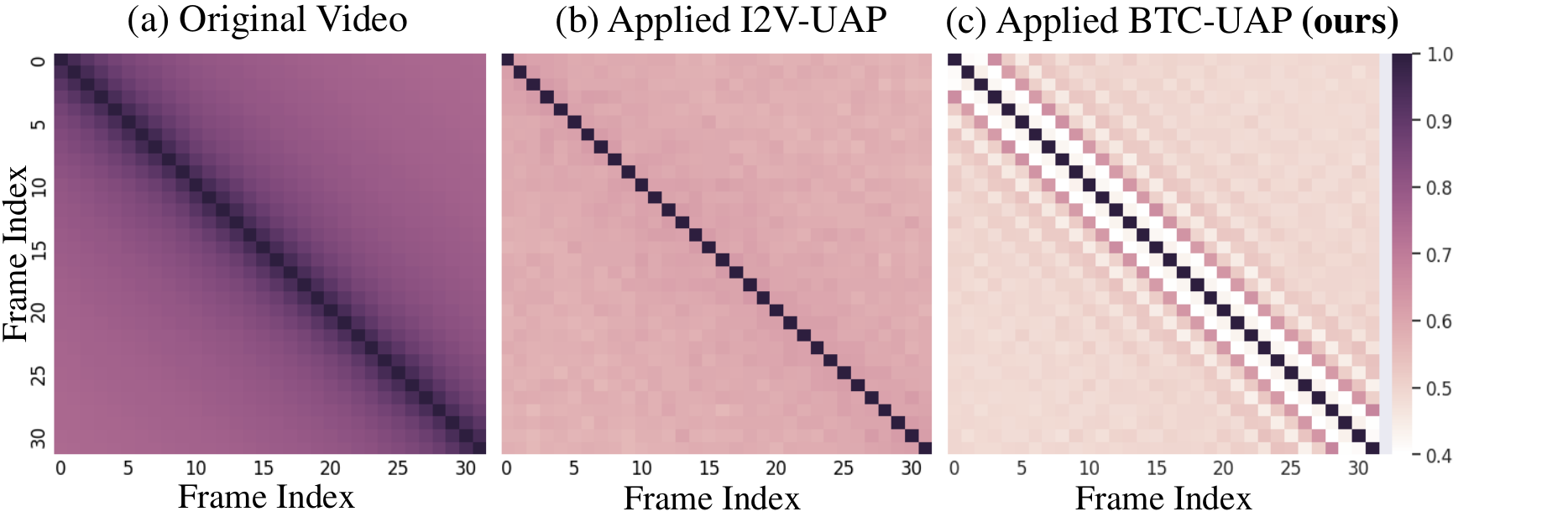}
         \vspace{-0.3cm}
         \caption{
        \textbf{The feature similarity of frames within videos.}
    This heatmap shows the average feature similarity between frames in the UCF-101 dataset, with brighter colors indicating lower levels of similarity.
        }
        \label{cos_sim_img}
        \vspace{-0.0cm}
\end{figure}

\section{Related Work}

\subsection{Adversarial Attacks}

Deep learning models are effective in computer vision tasks, but they can be easily fooled by adding imperceptible noise, which is known as adversarial perturbations.
The adversarial perturbation is added to the original data to create an adversarial example, 
and using this adversarial example to attack a deep learning model is called an adversarial attack.

\subsection{Image Classification Attacks}

As studies on adversarial attacks began with tricking image classification models, various image classification attack methods have been developed \cite{goodfellow2015explaining, kurakin2016adversarial, dong2018boosting, xie2019improving, dong2019evading, lin2020nesterov, byun2022improving, son2022adaptive}. In the first stage, white-box image-specific adversarial attack methods were introduced. Fast Gradient Sign Method (FGSM) \cite{goodfellow2015explaining} creates adversarial examples by updating an input image with its gradient calculated to increase the classification loss. FGSM evolved into an iterative method called Iterative Fast Gradient Sign Method (I-FGSM) \cite{kurakin2016adversarial}. I-FGSM iteratively updates the input image with its gradients calculated in the same way as FGSM. Then, Momentum Iterative Fast Gradient Sign Method (MI-FGSM) \cite{dong2018boosting} achieved better performance by integrating momentum during the iterative updates of I-FGSM.

Afterward, the transfer-based black-box attack methods have emerged. Diverse Input (DI) method \cite{xie2019improving} increases the transferability of adversarial examples by performing random resizing and random padding to input images at each iteration. Translation-Invariant (TI) method \cite{dong2019evading} uses multiple translated images to generate an adversarial perturbation, rather than using a single input image. 
They efficiently approximate this process by applying a convolutional operation with a kernel to the gradient obtained from a single input image without any translation.
Scale-Invariant (SI) attack method \cite{lin2020nesterov} improves the transferability of adversarial examples by using a scaled copy of the input image to compute the gradient. 

\cite{moosavi2017universal} showed the existence of a single adversarial perturbation that can fool image classifier models when added to any input images. This single perturbation is called a Universal Adversarial Perturbation (UAP). There are many studies on UAP designed for deep learning models that deal with images \cite{moosavi2017universal, mopuri2017fast, hayes2018learning, khrulkov2018art, mopuri2018generalizable, liu2019universal, zhang2020cd, li2022learning, zhang2021data}.  

\subsection{Video Classification Attacks}

There are several methods to create UAPs for video classification models \cite{wei2019sparse, hwang2021just, xie2022universal, li2018stealthy}. 
\cite{li2018stealthy} trains a Generative Adversarial Network (GAN) to generate UAPs, and \cite{xie2022universal} optimizes a noise generator to create a UAP.
\cite{wei2019sparse} and \cite{hwang2021just} are optimization-based white-box UAPs.
In white-box settings, \cite{wei2019sparse} introduced an optimization-based algorithm for generating adversarial perturbations on the whole video, specifically on LSTM-based models. They proposed to regularization to concentrate perturbations on key frames.
Similarly, a one-frame attack \cite{hwang2021just} only adds adversarial noise to one selected video frame. 
The researchers choose a vulnerable frame and perturbed it using the I-FGSM attack method. 
Similar to \cite{hwang2021just}, there are other key frame selection attack methods \cite{wei2020heuristic, xu2022sparse, du2022sparse} for both white-box and black-box settings.

In black-box settings, there are query-based video classification attacks \cite{jiang2019black, li2021adversarial, zhang2020motion, wei2022adaptive} and transfer-based video classification attacks \cite{TTwei2022boosting, I2Vwei2022cross}, similar to image classification attacks. 
\cite{TTwei2022boosting} introduced a method called TT (Temporal Translation) to enhance the transferability of video adversarial examples. They prevent overfitting the source model by optimizing over a set of video clips that have been translated in time for each video.
 I2V (Images to Videos) method \cite{I2Vwei2022cross} achieved better transferability without relying on video models. 
 I2V minimizes the similarity between the features of the original video frames and the adversarial video frames obtained by the ImageNet pre-trained image model. 
 These perturbations optimized with the image model applied to the videos to attack video models.
 Both previous works (TT and I2V) have significantly improved transferability, but they have the limitation of requiring optimization for each individual video, which is not the case for UAP.

\section{Methodology}

\begin{algorithm}[t]
    \algnewcommand\algorithmicinput{\textbf{Input:}}
    \algnewcommand\Input{\item[\algorithmicinput]}
    \algnewcommand\algorithmicoutput{\textbf{Output:}}
    \algnewcommand\Output{\item[\algorithmicoutput]}
    \caption{\hspace{0.1cm} \hsuap Attack Method} 
    \label{alg1}
    \begin{flushleft}
    
    \textbf{Input\hspace{0.76cm}:} 
            Image dataset $X\subset \mathbb{R}^{C \times H\times W}$,\\ 
            \hspace{1.76cm} image classification model $f(\cdot)$\\
    
    \textbf{Parameter:} 
            Perturbation budget $\epsilon$, \\
            \hspace{1.76cm} number of layer $l$, step size $\alpha$, \\
            \hspace{1.76cm} number of frames of \hsuap $N$,\\
            \hspace{1.76cm} number of random noises $K$,\\
            \hspace{1.76cm} set of temporal distance of neighbors $J$\\
            
        \textbf{Output\hspace{0.51cm}:} 
            \hsuap $\delta_N \in \mathbb{R}^{N \times C \times H\times W}$
            
\end{flushleft}
	\begin{algorithmic}[1]
            \State Initialize $n \leftarrow 1$
            \State Initialize all $\delta_N$ elements to $\frac{0.01}{255}$
    	\For{$x\in X$} 
                \State \textbf{$\triangleright$ Compute \hsnospace-Loss (\ref{eqn:total_loss}) with $l, J, K \text{ and }f$:} 
                \State $loss = \mathcal{L}_{BTC}(x, n, \delta_N)$
                
                \State \textbf{$\triangleright$ Update $\delta_N^n \in \delta_N$ by Adam optimizer:}
                \State $\delta_N^n \leftarrow Adam( loss, \alpha)$
    
                \State $\delta_N^n \leftarrow clip_{\epsilon}(\delta_N^n)$ 
                \State $n \leftarrow n+1$ 
                \If{$n > N$}
                    \State $ n\leftarrow 1$
                \EndIf
    	\EndFor
        \State \Return $\delta_N = \{\delta_N^1, ..., \delta_N^N\}$. \
	\end{algorithmic} 
\end{algorithm}

In this section, we describe the \hsfull method for generating the \hsuap using an image classification model that takes images or video frames as input.
This approach does not require any prior knowledge about the target video data or model and can fool the video model into producing an incorrect prediction.

\subsection{Problem Definition}
We consider a video $V\in\mathbb{R}^{T\times C \times H\times W}$ and aim to generate an adversarial video $V^{adv}$ by adding a \hsuap $\delta_N\in\mathbb{R}^{N\times C \times H\times W}$ to $V$.
Here, $T$, $C$, $H$, $W$, and $N$ denote the frames of the video, channels, height, width, and frames of the UAP, respectively.
To represent each frame of the $\delta_N$, we use $\delta_N^n\in\mathbb{R}^{C \times H\times W}$, where $n =1,..., N$ is the frame index.
To ensure the imperceptibility of the perturbation, we constrain $\delta_N$ to have an $l_{\infty}$-norm, as in previous works \cite{I2Vwei2022cross, hwang2021just}.

The value of $N$ is either less than or equal to $T$, and if $N < T$, we repeat the UAP in the frame dimension until it covers all $T$ frames of the video.
We define the repeated UAPs as $\delta_T \in \mathbb{R}^{T\times C \times H\times W}$, where $\delta_T = \{\delta_T^1, ..., \delta_T^T\}$ is obtained by repeating the original UAP $\delta_N = \{\delta_N^1, ...,  \delta_N^N\}$ in the frame dimension until it covers entire $T$ frames of the video.
We can represent this operation as follows:
\begin{align}\label{eqn:repeat}
\delta_T = Repeat(\delta_N) = \{\underbrace{\delta_N^{1}, ..., \delta_N^{N}, \delta_N^{1},..., \delta_N^{N}, \delta_N^{1}, ...}_{\text{repeated until it covers } T \text{ frames}}\}.
\end{align}

Let $g(\cdot)$ be a video recognition model, and $y$ be the true label of $V$.
Our goal is to find a perturbation $\delta_N$ that misleads the video model's prediction:
\vspace{-1mm} 
\begin{align}\label{eqn:attack}
 g(V + \delta_T) \neq y, \quad s.t. \quad || \delta_T||_\infty \leq \epsilon .
\end{align}
To achieve this, we optimize $\delta_N$ with $f(\cdot)$, which represents a image classification model. 

\subsection{Feature Similarity Analysis of Video Frames}

In this section, we measure the average similarity of features obtained between video frames in the dataset.
Since feature maps represent characteristics of an image, we use them to compare the similarity between frames. 
Figure \ref{cos_sim_img} represents the feature similarity between two frames of the videos.
For example, the diagonal represents the similarity between identical frames, so it always has the value of 1.
To obtain the value, we input each frame of the video to an image model and measured the similarity $Sim$ at a specific feature level using cosine similarity. 
The similarity between vectors $\mathbf{x}_1$ and $\mathbf{x}_2$ is expressed as follows:
\vspace{-0mm} 
\begin{align}\label{eqn:cossim}
Sim(\mathbf{x}_1, \mathbf{x}_2) = \frac{\mathbf{x}_1 \cdot \mathbf{x}_2}{\left\lVert\mathbf{x}_1\right\rVert \left\lVert\mathbf{x}_2\right\rVert}.
\end{align}

To represent each frame of the video, we use $V^t\in\mathbb{R}^{C \times H\times W}$, where $t \in T$ is the frame index. We extract the feature map $F(\cdot)$ from a specific layer $l$ of an image classification model $f(V^t)$ and denote this feature map by $F_{l}(V^{t})$. We visualize the similarity of frames within an original video $V$ in Figure \ref{cos_sim_img}-(a). We observe that the original videos tend to have high levels of similarity between consecutive frames.

Furthermore, we extend the non-UAP I2V method \cite{I2Vwei2022cross} to create an I2V-UAP. To make I2V universal, we optimize one perturbation for multiple videos within the dataset. To observe the effects of UAPs on the feature maps, we create an adversarial example $V_{\text{adv}}^{t}$ by adding the UAP $\delta_T^t$ to the original frame $V^{t}$ and extract the feature maps $F_{l}(V_{\text{adv}}^{t})$ in the same way as for the original frame. Applying the I2V-UAP shown in Figure \ref{cos_sim_img}-(b) results in a reduction in similarity across all frames.

We further observe that adversarial videos disrupt the high similarity pattern of consecutive frames in the original videos.
Based on this observation, we propose the \hs method to generate adversarial videos with opposite patterns to the original videos.
Details of our method can be found in Section \ref{subsec:hs-method}.
Our proposed \hsuapnospace, as shown in Figure \ref{cos_sim_img}-(c), generates a completely opposite pattern of similarity to the original video, with neighboring frames having low similarity.
As we intentionally make the features of consecutive frames less similar to each other, the overall similarity between frames decreases when compared to the original video.
These results indicate that neighboring frames are recognized as different images by image models.
These effects are contrary to the original characteristics of the video, and our experiments in Section \ref{sec:exp} demonstrates that \hsuap effectively confuses video models.

\subsection{\hsfull Method}
\label{subsec:hs-method}

In this section, we focus on \hsfull method and discuss how to optimize the \hsuap using image data and models.
Let $x\in\mathbb{R}^{C \times H\times W}$ be an image, which can be a frame of a video $V^t\in\mathbb{R}^{C \times H\times W}$.
Our goal is to find a universal adversarial perturbation $\delta_N^n \in \delta_N$ using images.
The overall optimization process is described in Algorithm \ref{alg1}. 

\textbf{Adversarial Loss.} 
Feature maps represent characteristics and patterns of an image, which can be used to create adversarial examples. Therefore, decreasing the similarity between the feature representations $F(\cdot)$ of original images $x$ and adversarial examples $x_{adv}^n = x+ \delta_N^n$ will result in the UAP causing confusion in the information of the original image.
To ensure that the \hsuap is effective against other data and prevent overfitting to the training dataset, we propose \textit{Feature Diversity} method with a total of $K$ random noises. This involves adding a random noise $\eta_k \in [-\epsilon, \epsilon]^{C \times H\times W}$ to each $x$ to increase diversity to avoid overfitting. This simple method is highly effective in improving the performance of the UAP framework.
The adversarial loss can be expressed mathematically as follows:

\vspace{-0mm} 
\begin{align}\label{eqn:adversarial_loss}
\mathcal{L}_{adv}(x,n,\delta_N) = \sum_{k=1}^{K} Sim(F_l(x + \eta_k), F_l(x_{adv}^n)).
\end{align}

\textbf{Temporal Similarity Loss.}
Our approach presents a novel solution to the issue that image models are unable to fully consider the temporal dimension, in contrast to video models. 
Our goal is to minimize the similarity between neighboring frames in videos using the optimized $\delta_N$. 
To successfully deceive a video model, we introduce confusion in the temporal domain through the use of $f(\cdot)$, by decreasing similarity between the neighboring frames.
To achieve this, we generate the adversarial image $x_{n+j}^\text{adv} = x + \delta^{n+j}_{N}$ and then treat the sequence of adversarial images as a pseudo video.
Here, $j \in J$ and $J$ represents the set of temporal distances of neighbors, such as $J = \{-2,-1, 1, 2\}$.

To reduces the similarity between $x_{adv}^{n}$ and $x_{adv}^{n+j}$, we extract feature of adversarial images $F(x_{adv})$ using the image model $f(\cdot)$ and calculate the similarity between them, following Eq.\ref{eqn:cossim}.
The temporal similarity loss can effectively cause confusion in the temporal information when the perturbations $\delta_N$ is added to video along the temporal axis. 
This temporal similarity loss can be expressed mathematically as follows:

\vspace{-0mm}
\begin{align}\label{eqn:temporal_similarity_loss}
\mathcal{L}_{temp}(x,n, \delta_N) = \sum_{j \in J} Sim\left(F_l(x_{adv}^n), F_l(x_{adv}^{n+j})\right).
\end{align}
Compared to previous approaches, our method allows us to effectively minimize the temporal similarity between perturbed frames, enabling us to produce more robust adversarial examples.

By considering both adversarial and temporal aspects using image-based approaches, the proposed \hsuap can effectively perturb both types of information and successfully attack video models.
To optimize $\delta_N$, we utilized a \hsfull Loss that is the sum of the adversarial loss and temporal similarity loss, mathematically represented as follows:

\vspace{-3mm}
\begin{align}
\mathcal{L}_{BTC}(x,n,\delta_N) =\mathcal{L}_{adv} + \mathcal{L}_{temp}.
\end{align}

Finally, we can get optimized \hsuap $\delta_N^{n*}$ by minimizing \hsnospace-Loss with $l, J, K \text{ and }f$:
\vspace{-0mm}
\begin{align}\label{eqn:total_loss}
\delta_N^{n*} = arg\min_{\delta_N^n} \mathcal{L}_{BTC}(x, n, \delta_N).
\end{align}
\section{Experiment} \label{sec:exp}
\begin{table}[!t]
\small
\centering
\resizebox{\columnwidth}{!}{%
\setlength\tabcolsep{1.5pt}
\begin{tabular}{cccccccc}
\toprule
                          & \multicolumn{7}{c}{Networks}                                                                                                                                                                                    \\
\multirow{-2}{*}{Dataset} & SF-101                      & SF-50                       & TPN-50                      & TPN-101                     & NL-50                      & NL-101                     & AVG.                        \\ \hline
UCF-101                   & {\color[HTML]{333333} 90.2} & {\color[HTML]{333333} 91.7} & {\color[HTML]{333333} 91.7} & {\color[HTML]{333333} 93.6} & {\color[HTML]{333333} 86.9} & {\color[HTML]{333333} 88.4} & {\color[HTML]{333333} 90.4} \\
Kinetics                  & {\color[HTML]{333333} 69.8} & {\color[HTML]{333333} 71.0} & {\color[HTML]{333333} 73.9} & {\color[HTML]{333333} 75.0} & {\color[HTML]{333333} 69.5} & {\color[HTML]{333333} 69.5} & 71.4                        \\ \bottomrule
\end{tabular}
}
\caption{Clean Accuracy}
\label{tab:clean}
\end{table}
\definecolor{Gray}{gray}{0.92}

\begin{table*}[!t]
\centering
\resizebox{0.9\textwidth}{!}{%
\begin{tabular}{cccccccccc}
\toprule
                                                                           &                                                                                &                          & \multicolumn{6}{c}{Target Models}                                                                                                                                        &                        \\
                                                                           &                                                                                &                          & SF-50          & SF-101                                & TPN-50         & TPN-101                               & NL-50         & NL-101                               &                        \\
\multirow{-3}{*}{\begin{tabular}[c]{@{}c@{}}Source\\ Dataset\end{tabular}} & \multirow{-3}{*}{\begin{tabular}[c]{@{}c@{}}Source\\ Models\end{tabular}}      & \multirow{-3}{*}{Attack} & (Kinetics)        & (Kinetics)                               & (Kinetics)        & (Kinetics)                               & (Kinetics)        & (Kinetics)                               & \multirow{-3}{*}{AVG.} \\ \hline
                                                                           &                                                                                & All-UAP                  & 76.12          & {\color[HTML]{808080} \textit{98.59}} & 49.76          & 45.90                                 & 48.41          & 48.76                                 & 61.26                  \\
                                                                           & \multirow{-2}{*}{\begin{tabular}[c]{@{}c@{}}SF-101\\ (UCF-101)\end{tabular}}   & TT-UAP                   & 75.92          & {\color[HTML]{808080} \textit{98.74}} & 66.59          & 58.97                                 & 62.55          & 62.55                                 & 70.88         \\ \cline{2-10} 
                                                                           &                                                                                & All-UAP                  & 58.81          & 58.29                                 & 62.91          & {\color[HTML]{808080} \textit{79.95}} & 51.29          & 48.49                                 & 59.95                  \\
                                                                           & \multirow{-2}{*}{\begin{tabular}[c]{@{}c@{}}TPN-101\\ (UCF-101)\end{tabular}}  & TT-UAP                   & 53.99          & 53.16                                 & 48.85          & {\color[HTML]{808080} \textit{44.68}} & 43.86          & 41.62                                 & 47.69                  \\ \cline{2-10} 
                                                                           &                                                                                & All-UAP                  & 52.75          & 50.21                                 & 39.78          & 37.93                                 & \textbf{69.80} & {\color[HTML]{808080} \textit{81.32}} & 55.30                  \\
                                                                           & \multirow{-2}{*}{\begin{tabular}[c]{@{}c@{}}NL-101\\ (UCF-101)\end{tabular}}  & TT-UAP                   & 59.69          & 60.21                                 & 60.60          & 56.20                                 & 67.41          & {\color[HTML]{808080} \textit{80.52}} & 64.10         \\ \cline{2-10} 
                                                                           &                                                                                & I2V-UAP                  & 66.29          & 62.15                                 & 82.40          & 71.78                                 & 51.85          & 50.18                                 & 64.11                  \\
\multirow{-8}{*}{UCF-101}                                                  & \multirow{-2}{*}{\begin{tabular}[c]{@{}c@{}}Res-101\\ (ImageNet)\end{tabular}} & \cellcolor{Gray} BTC-UAP                   & \cellcolor{Gray} 82.49          & \cellcolor{Gray} 77.30                                 & \cellcolor{Gray} 93.35          & \cellcolor{Gray} \textbf{86.62}                        & \cellcolor{Gray} 67.40          & \cellcolor{Gray} \textbf{67.83}                                 & \cellcolor{Gray} 79.16                  \\ \hline
                                                                           &                                                                                & I2V-UAP                  & 69.31          & 64.37                                 & 83.85          & 73.64                                 & 53.26          & 50.97                                 & 65.90                  \\
\multirow{-2}{*}{ImageNet}                                                 & \multirow{-2}{*}{\begin{tabular}[c]{@{}c@{}}Res-101\\ (ImageNet)\end{tabular}} & \cellcolor{Gray} BTC-UAP                   & \cellcolor{Gray} \textbf{83.26} & \cellcolor{Gray} \textbf{77.63}                        & \cellcolor{Gray} \textbf{93.49} & \cellcolor{Gray} 86.23                        & \cellcolor{Gray} 68.27          & \cellcolor{Gray}66.97                        & \cellcolor{Gray} \textbf{79.31}         \\ \bottomrule
\end{tabular}
}
\caption{\textbf{Comparison with UAPs generated on video models.}
UAPs are optimized on the source datasets UCF-101 and ImageNet, respectively. The generated UAPs are \textit{repeated} and added to Kinetics-400 videos until they cover the entire video. The bold numbers indicate the highest attack success rates (\%) in each column. The gray color represents the \textit{white-box setting}, where the source and target models are identical.
}
\label{tab:video-kinetics}
\end{table*}

\begin{table*}[!t]
\centering
\resizebox{0.9\textwidth}{!}{%
\begin{tabular}{cccccccccc}
\toprule
                                                                           &                                                                                &                          & \multicolumn{6}{c}{Target Models}                                                                                                                                        &                        \\
                                                                           &                                                                                &                          & SF-50          & SF-101                                & TPN-50         & TPN-101                               & NL-50         & NL-101                               &                        \\
\multirow{-3}{*}{\begin{tabular}[c]{@{}c@{}}Source\\ Dataset\end{tabular}} & \multirow{-3}{*}{\begin{tabular}[c]{@{}c@{}}Source\\ Models\end{tabular}}      & \multirow{-3}{*}{Attack} & (UCF-101)        & (UCF-101)                               & (UCF-101)        & (UCF-101)                               & (UCF-101)        & (UCF-101)                               & \multirow{-3}{*}{AVG.} \\ \hline
                                                                           &                                                                                & All-UAP                  & 48.74          & {\color[HTML]{808080} \textit{98.93}} & 13.42          & 8.62                                  & 21.18          & 14.14                                 & 34.17                  \\
                                                                           & \multirow{-2}{*}{\begin{tabular}[c]{@{}c@{}}SF-101\\ (UCF-101)\end{tabular}}   & TT-UAP                   & 43.28          & {\color[HTML]{808080} \textit{96.81}} & 23.25          & 17.01                                 & 38.38          & 50.46                                 & 44.86                  \\ \cline{2-10} 
                                                                           &                                                                                & All-UAP                  & 17.62          & 12.88                                 & 19.93          & {\color[HTML]{808080} \textit{94.54}} & 26.83          & 27.00                                 & 33.13                  \\
                                                                           & \multirow{-2}{*}{\begin{tabular}[c]{@{}c@{}}TPN-101\\ (UCF-101)\end{tabular}}  & TT-UAP                   & 14.41          & 8.11                                  & 9.88           & {\color[HTML]{808080} \textit{6.88}}  & 17.01          & 15.48                                 & 11.96                  \\ \cline{2-10} 
                                                                           &                                                                                & All-UAP                  & 19.20          & 10.23                                 & 8.17           & 5.52                                  & \textbf{56.19} & {\color[HTML]{808080} \textit{97.96}} & 32.88                  \\
                                                                           & \multirow{-2}{*}{\begin{tabular}[c]{@{}c@{}}NL-101\\ (UCF-101)\end{tabular}}  & TT-UAP                   & 20.84          & 18.02                                 & 23.73          & 21.34                                 & 51.47          & {\color[HTML]{808080} \textit{96.79}} & 38.70                  \\ \cline{2-10} 
                                                                           &                                                                                & I2V-UAP                  & 24.24          & 16.71                                 & 39.21          & 27.02                                 & 24.18          & 40.01                                 & 28.56                  \\
\multirow{-8}{*}{UCF-101}                                                  & \multirow{-2}{*}{\begin{tabular}[c]{@{}c@{}}Res-101\\ (ImageNet)\end{tabular}} & \cellcolor{Gray} BTC-UAP                   & \cellcolor{Gray} 47.78          & \cellcolor{Gray} 35.62                                 & \cellcolor{Gray} 64.43          & \cellcolor{Gray} 46.55                                 & \cellcolor{Gray} 50.43          & \cellcolor{Gray} 61.89                                 & \cellcolor{Gray} 51.12                  \\ \hline
                                                                           &                                                                                & I2V-UAP                  & 25.60          & 18.45                                 & 42.55          & 29.16                                 & 25.82          & 41.27                                 & 30.48                  \\
\multirow{-2}{*}{ImageNet}                                                 & \multirow{-2}{*}{\begin{tabular}[c]{@{}c@{}}Res-101\\ (ImageNet)\end{tabular}} & \cellcolor{Gray} BTC-UAP                   & \cellcolor{Gray} \textbf{49.01} & \cellcolor{Gray} \textbf{36.98}                        & \cellcolor{Gray} \textbf{65.37} & \cellcolor{Gray} \textbf{47.67}                        & \cellcolor{Gray} 49.41          & \cellcolor{Gray} \textbf{63.34}                        & \cellcolor{Gray} \textbf{51.96}         \\ \bottomrule
\end{tabular}

}
\caption{\textbf{Comparison with UAPs generated on video models.}
UAPs are optimized on the source datasets UCF-101 and ImageNet, respectively. Adversarial videos are generated by adding UAPs to UCF-101 videos. The bold numbers indicate the highest attack success rates (\%) in each column for the UCF-101 dataset. The gray color represents the \textit{white-box} setting, where the source model and target model are identical.
}
\vspace{-0.3cm}
\label{tab:video-ucf}
\end{table*}


\definecolor{Gray}{gray}{0.92}

\begin{table*}[!t]
\small
\centering
\resizebox{0.8\textwidth}{!}{%
\begin{tabular}{ccccccccc}
\toprule
\multirow{3}{*}{\begin{tabular}[c]{@{}c@{}}Source\\ Models\end{tabular}}      & \multirow{3}{*}{Attack} & \multicolumn{6}{c}{Target Models}                                                                   & \multirow{3}{*}{AVG.} \\
                                                                              &                         & SF-50          & SF-101         & TPN-50         & TPN-101        & NL-50         & NL-101        &                       \\
                                                                              &                         & (Kinetics)       & (Kinetics)       & (Kinetics)       & (Kinetics)       & (Kinetics)       & (Kinetics)       &                       \\ \hline
\multirow{6}{*}{\begin{tabular}[c]{@{}c@{}}Res-101\\ (ImageNet)\end{tabular}} & MI-UAP                  & 61.42          & 57.46          & 58.02          & 54.43          & 49.36          & 47.37          & 54.36                 \\
                                                                              & DI-UAP                  & 60.55          & 56.63          & 59.26          & 58.66          & 51.64          & 48.91          & 55.94                 \\
                                                                              & TI-UAP                  & 60.56          & 56.63          & 59.26          & 58.64          & 56.57          & 54.41          & 57.68                 \\
                                                                              & SI-UAP                  & 65.98          & 64.61          & 71.26          & 70.13          & 55.72          & 55.30          & 63.83                 \\
                                                                              & I2V-UAP                 & 69.31          & 64.37          & 83.85          & 73.64          & 53.26          & 50.97          & 65.90                 \\
                                                                              & \cellcolor{Gray} BTC-UAP                  & \cellcolor{Gray} \textbf{83.26} & \cellcolor{Gray} \textbf{77.63} & \cellcolor{Gray} \textbf{93.49} & \cellcolor{Gray} \textbf{86.23} & \cellcolor{Gray} \textbf{68.27} & \cellcolor{Gray} \textbf{66.97} & \cellcolor{Gray} \textbf{79.31}        \\ \hline
\multirow{6}{*}{\begin{tabular}[c]{@{}c@{}}VGG16\\ (ImageNet)\end{tabular}}   & MI-UAP                  & 54.42          & 52.56          & 49.53          & 45.43          & 45.64          & 43.44          & 49.11                 \\
                                                                              & DI-UAP                  & 57.96          & 55.47          & 51.97          & 48.22          & 49.25          & 45.96          & 51.47                 \\
                                                                              & TI-UAP                  & 57.95          & 55.45          & 51.95          & 48.21          & 55.51          & 52.63          & 53.61                 \\
                                                                              & SI-UAP                  & 59.63          & 58.29          & 56.22          & 51.25          & 50.19          & 48.67          & 54.04                 \\
                                                                              & I2V-UAP                 & 57.82          & 52.73          & 60.40          & 54.90          & 44.62          & 41.92          & 52.06                 \\
                                                                              & \cellcolor{Gray} BTC-UAP                  & \cellcolor{Gray} \textbf{74.25} & \cellcolor{Gray} \textbf{75.05} & \cellcolor{Gray} \textbf{82.97} & \cellcolor{Gray} \textbf{75.68} & \cellcolor{Gray} \textbf{68.59} & \cellcolor{Gray} \textbf{63.93} & \cellcolor{Gray} \textbf{73.41}        \\ \hline
\multirow{6}{*}{\begin{tabular}[c]{@{}c@{}}Squeeze\\ (ImageNet)\end{tabular}} & MI-UAP                  & 53.91          & 52.92          & 52.22          & 49.55          & 45.64          & 43.44          & 49.61                 \\
                                                                              & DI-UAP                  & 54.61          & 53.42          & 52.22          & 49.24          & 45.24          & 43.61          & 49.72                 \\
                                                                              & TI-UAP                  & 66.25          & 66.05          & 59.55          & 54.79          & 57.52          & \textbf{55.06} & 59.87                 \\
                                                                              & SI-UAP                  & 58.36          & 58.40          & 54.51          & 49.86          & 47.70          & 45.70          & 52.42                 \\
                                                                              & I2V-UAP                 & 66.20          & 63.40          & 66.50          & 58.94          & 54.30          & 50.60          & 59.99                 \\
                                                                              & \cellcolor{Gray} BTC-UAP                  & \cellcolor{Gray} \textbf{70.71} & \cellcolor{Gray} \textbf{68.14} & \cellcolor{Gray} \textbf{71.29} & \cellcolor{Gray} \textbf{65.25} & \cellcolor{Gray} \textbf{58.88} & \cellcolor{Gray} 54.72          & \cellcolor{Gray} \textbf{64.83}        \\ \hline
\end{tabular}

}
\caption{\textbf{Attack success rates (\%) of UAPs generated on image models using image data.}
UAPs are optimized on ImageNet and adversarial videos are generated by adding UAPs to Kinetics-400 videos. 
The generated UAPs are \textit{repeated} and added to Kinetics-400 videos until they cover the
entire video.
The bold numbers indicate the highest attack success rate among attack methods.
}
\label{tab:image-kinetics}
\end{table*}

\subsection{Experiment Settings}
We evaluate the Attack Success Rates (ASR) of UAPs in the following settings.
The ASR indicates the rate at which the target model misclassifies the adversarial examples into the wrong label.
A higher ASR indicates that the UAPs achieve higher transferability.

\textbf{Datasets.}
We refer to the data used to generate UAPs as the source data, and the data where the UAP is added to create adversarial examples as the target data.
We conducted experiments using various datasets. 
ImageNet \cite{deng2009imagenet} is a large image dataset with 1,000 classes. 
We used the ImageNet train set as source data, selecting 10 images per one class. 
UCF-101 \cite{soomro2012ucf101} and Kinetics-400 \cite{kay2017kinetics} are video classification datasets that label human action categories.
UCF-101 has 13,320 videos with 101 action classes, and Kinetics-400 has 650,000 videos with 400 classes. 
We used the UCF-101 test set and Kinetics-400 validation set. 
For Kinetics-400, we randomly chose 5 videos per a class.

\textbf{Models.}
We used three pre-trained image models on the ImageNet dataset: ResNet101 (Res-101) \cite{he2016deep}, SqueezeNet (Squeeze) \cite{iandola2016squeezenet}, and VGG16 \cite{simonyan2014very}.
These models were used as source models to generate adversarial examples. 
We used six different video models: SlowFast-50 (SF-50), SlowFast-101 (SF-101)\cite{feichtenhofer2019slowfast}, Temporal Pyramid Network-50 (TPN-50), Temporal Pyramid Network-101 (TPN-101) \cite{yang2020temporal}, NonLocal-50 (NL-50), and NonLocal-101 (NL-101) \cite{wang2018non}. 
Each six models trained on UCF-101\footnote{We used the trained models from https://github.com/zhipeng-wei/Image-to-Video-I2V-attack.} and Kinetics-400\footnote{We downloaded the pretrained models from gluon.} datasets, for a total of 12 video models are used to evaluate the performance.
UCF-101 models were tested on 32-frame videos, while Kinetics-400 models were tested on 64-frame videos.
Table \ref{tab:clean} shows the accuracy of the models on clean data.

\textbf{Baselines.}
There is no UAP framework for transfer-based video attacks using image datasets and models.
To compare performance, we adapted the cross-modal video attack method \cite{I2Vwei2022cross} and the transfer-based video attack method \cite{TTwei2022boosting} to the UAP scenario (I2V-UAP and TT-UAP).
ALL-UAP indicates UAP based on I-FGSM method \cite{kurakin2016adversarial}.
We also compared our method with image transfer-based attack methods, including MI \cite{dong2018boosting}, DI \cite{dong2019evading}, TI \cite{xie2019improving}, and SI \cite{lin2020nesterov} in Table \ref{tab:image-kinetics}.

\textbf{Hyperparameters.}
The perturbation budget $\epsilon$ was set to $16/255$, and the step size $\alpha$ was set to $0.004$.
We used the number of feature layers $l$ to optimize \hsuap and I2V-UAP, following the previous cross-modal attack \cite{I2Vwei2022cross}.
We randomly selected an image or one frame per a video for \hsuap optimization, and set the number of UAP frames $N$ to 32.
The number of random noise $K$ was set to 4 and the set of temporal distance $J$ to $\{ -2, -1, +1, +2\}$.

In Section \ref{subsec:abl}, we show how we selected the hyperparameters for \hsuapnospace.
The implementation details for other baselines are in the supplementary material.
\begin{figure*}[t!]
     \centering
         \centering
         \includegraphics[width=0.99
         \textwidth,trim={0.0cm 0cm 0cm 0cm},clip] {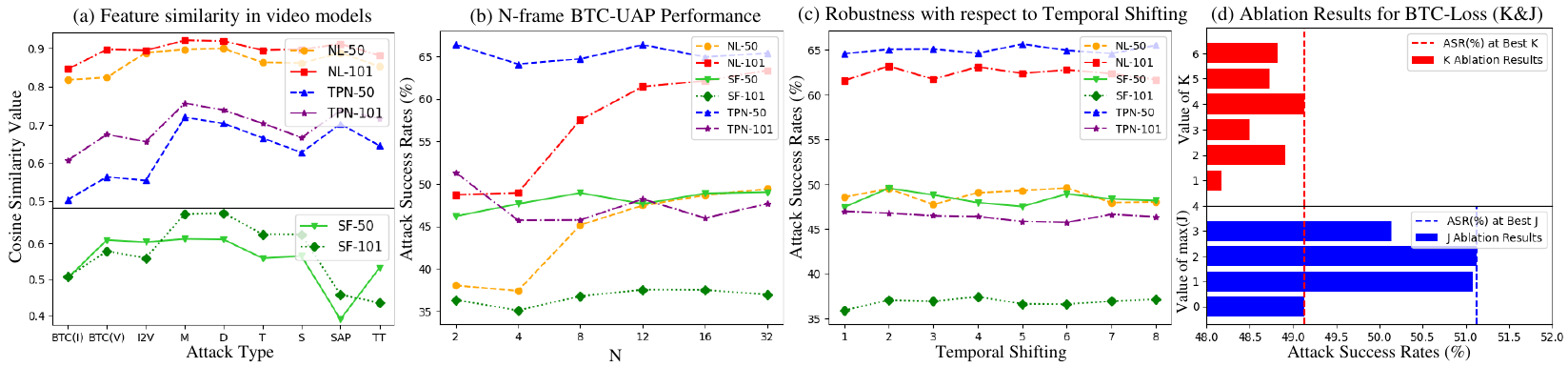}
         \vspace{0.0cm}
         \caption{
        \textbf{Analysis and Ablation Results.} 
Results demonstrate that the proposed Breaking Temporal Consistency method leads to superior robustness against perturbations for videos.
        }
        \label{fig_D1}
        \vspace{-0.3cm}
\end{figure*}


\subsection{Experimental Results}

\subsubsection{Comparison with Video-based attack method}
We evaluated the transferability of UAPs optimized on UCF-101 in Tables \ref{tab:video-kinetics} and \ref{tab:video-ucf}.
Table \ref{tab:video-kinetics} shows the performance of UAPs on each target model trained on Kinetics-400, evaluated by adding the UAPs to Kinetics-400 videos.
In Table \ref{tab:video-ucf}, we divided the UCF-101 dataset into two groups and evaluated methods on the unseen group.
The gray color in the tables represents the \textit{white-box setting}, where the target model is used as the source model during UAP generation. 
Please note that our method aims to transfer the UAPs generated from image models to video models for cross-modal attacks, which cannot be conducted under the white-box setting. 
Excluding the white-box evaluation, \hsuap achieves the highest transferability in most cases.
For example, in Table \ref{tab:video-kinetics}, \hsuap(Res-101, ImageNet) achieved the highest average ASR of 79.31\%, compare to All-UAP(SF-101, UCF-101) with 61.26\% and TT-UAP(SF-101, UCF-101) with 70.88\%.

When compared to UAPs optimized using videos as the source data, the performance of \hsuap generated on image data is comparable or even better. This demonstrates that our \hsfull method can effectively consider temporal information, even without video models or data, and achieve superior performance compared to I2V-UAP.
Furthermore, in Table \ref{tab:video-kinetics}, the generated UAP was optimized for 32 frames, while the evaluation on Kinetics-400 was conducted on 64 frames.
Therefore, we \textit{repeated} the UAP without optimizing it for 64 frames to generate universal adversarial perturbations for Kinetics-400, following Eq.\ref{eqn:repeat}. 
Despite the challenging condition of evaluating on 64 frames while the generated UAP was optimized for 32 frames, \hsuap still achieves high transferability in attacking video classification models. This demonstrates that our method is effective even in complete black-box situations, such as evaluating on an unseen video model with a different number of video frames.

\subsubsection{Comparison with Image-based attack method}

We conducted experiments to evaluate the transferability of UAPs generated using image data and models. 
Table \ref{tab:image-kinetics} shows the ASR of adversarial videos, where UAP is optimized on ImageNet using each rightmost image model. 
In this experiment, the 32-frame UAPs are repeatedly added to Kinetics-400 videos to create adversarial videos, following Eq.\ref{eqn:repeat}.
Compared to other methods, \hsuap achieved the highest average ASR and demonstrated good transferability.
For instance, in Table \ref{tab:image-kinetics}, the I2V-UAP has a total average ASR 60.31 \% on all cases, while \hsuap shows superior performance with 70.79 \%.
This result demonstrates that our proposed method effectively considers temporal information, resulting in the highest performance among image-based methods.

\subsection{Discussion}

In this section, we conducted experiments to demonstrate the effectiveness of our proposed \hsfull Loss for creating adversarial examples in video-based attacks. 
We applied the \hsuap generated using Res-101 and ImageNet data, to 32-frame UCF-101 videos and analyzed its effectiveness on six different models.

To analyze the performance of our proposed method, we compared the cosine similarity between the original and adversarial videos in the video model. 
Figure\ref{fig_D1}-(a) shows cosine similarity scores between the original and adversarial videos for each attack method under black-box settings. The bottom graph shows the same comparison under white-box settings. 
when all attack settings are black-box, our proposed method achieved the lowest similarity score among all the attack methods.
In the context of confusing video models, we found that \hsnospace(I), which is generated using image data, is more effective than \hsnospace(V), which is generated using video data.
The results show that \hsnospace(I) had a greater impact the UAPs generated with video models despite being generated using image models, highlighting its superior robustness.

To demonstrate the effectiveness of the proposed method with a small number of \hsuap frames, 
we applied the UAP iteratively with a small value of N, repeating a subset of N frames within the total of $T=32$ frames in the adversarial video.
We compared the results for N=2,4,8,12,16 and 32.
 Figure\ref{fig_D1}-(b) compares the performance of \hsuap with different numbers of N. 
Even when N=2, our proposed method exhibits comparable performance, demonstrating its effectiveness even with a small number of UAP frames.
We further demonstrated the shifting invariance of our proposed \hsuap by conducting experiments in which we shifted the UAP along the temporal axis from 1 to 8 frames. 
Figure\ref{fig_D1}-(c) demonstrates the shifting invariance of \hsuap by displaying attack success rates for different temporal shifts of the UAP frames. 
It showed that the attack success rate was consistent regardless of the temporal shifting.
These results demonstrated the \hsuap is robustness against temporal shifts and the effectiveness even with a small number of optimized frames.

\subsection{Ablation study}\label{subsec:abl}

In this section, we explore the effects of the most critical parameters, K and J, in our \hsnospace-method. Specifically, we investigate the impact of the number of random noise $K$ employed in the adversarial loss and the temporal distance of neighbors set $J$ utilized in the temporal similarity loss.
Figure \ref{fig_D1}-(d) shows that $K=4$ provided the best performance in terms of the adversarial loss. We then conducted experiments with different symmetric sets of $J$ while keeping $K$ fixed at 4. In the graph, please note that we represented the highest value among the set of $J$ on the y-axis for convenience. Our results showed that when the $max(J)=2$, the use of a set $J = \{-2,-1,1,2\}$ achieved the highest performance. 

Importantly, we observed that although the computations required for $K=6$ and $K=4$ with $J =\{-1,1\}$ were the same, the latter yielded significantly better performance. This demonstrates that reducing the similarity between frames was a more effective approach to improving performance than simply increasing computational resources.
\section{Conclusion}

In this paper, we proposed the \hsfull Method, which was the first to attack videos using only image models while considering temporal information.
Our method was designed to minimize the similarity between neighboring frames, by jointly optimizing adversarial and temporal similarity losses.
Specifically, by using adversarial loss, we reduced the similarity between original and adversarial examples, and by using temporal similarity loss, we reduced the similarity between UAPs.
\hsuap was both temporal shift invariant and length-agnostic.
Our extensive experiments on various datasets demonstrated the effectiveness of our proposed \hsuap.

\vfill\pagebreak

{\small
\bibliographystyle{ieee_fullname}
\bibliography{0_main}
}

\end{document}